\documentclass[letterpaper]{article}
\usepackage[preprint]{aaai2027}
\usepackage[hyphens]{url}
\usepackage{graphicx}
\usepackage{natbib}
\usepackage{caption}
\usepackage{amsmath,amssymb}
\usepackage{booktabs}
\usepackage{multirow}

\newcommand{\rankFirst}[1]{\textbf{#1}}
\newcommand{\rankSecond}[1]{\underline{#1}}
\newcommand{\rankThird}[1]{\textit{#1}}

\title{MotionStrata: Hierarchical Motion Latents for Compact Video Autoencoding}
\author{
Wenzhang Sun\equalcontrib\textsuperscript{\rm 1},
Huaize Liu\equalcontrib\textsuperscript{\rm 2,3},
Chunfeng Wang,\textsuperscript{\rm 1}
Biao Gong,\textsuperscript{\rm 4}
Hao Li,\textsuperscript{\rm 1}
Changqing Zou\textsuperscript{\rm 3,5}
}
\affiliations{
\textsuperscript{\rm 1}Li Auto \quad
\textsuperscript{\rm 2}University of Chinese Academy of Sciences \quad
\textsuperscript{\rm 3}Zhejiang Lab \\
\textsuperscript{\rm 4}Ant Group \quad
\textsuperscript{\rm 5}State Key Laboratory of CAD\&CG, Zhejiang University
}

\begin{document}
\maketitle
\begin{abstract}
First-frame-conditioned video autoencoders represent a clip with persistent content and a compact motion code. Although this removes much of the appearance redundancy, the remaining motion is typically compressed with a homogeneous latent geometry. Such representations use the same temporal support for broad scene evolution and fine-grained, frame-specific details. We introduce MotionStrata, which organizes a fixed motion budget into Global Motion and Detailed Motion. Temporally compressed Global queries summarize broad evolution, whereas frame-aligned Detailed queries preserve fine-grained structures whose configuration varies across frames. Frequency-guided routing and coarse-to-fine training establish this hierarchy without increasing motion dimensionality. Experiments show that MotionStrata maintains high reconstruction quality under aggressive compression and outperforms uniform and alternative grouped representations. Additional experiments evaluate hierarchical representation, downstream generation, and decoding cost. These results support hierarchical motion organization as a useful design principle for compact video autoencoding.
\end{abstract}

\section{Introduction}
\label{sec:intro}

Video representations must compress substantial temporal redundancy without losing the fine-grained structures that distinguish individual frames. Appearance can persist throughout a clip, while broad camera and body motion often evolves smoothly. At the same time, small objects, articulated parts, and deformation boundaries require precise frame-level representation. First-frame-conditioned autoencoders reduce appearance redundancy by anchoring persistent content to a reference frame and reserving a smaller latent for temporal variation~\cite{tian2024reducio,wang2024vidtwin}. Yet the remaining motion is usually encoded as a homogeneous stream.

Existing content--motion factorization primarily determines what can be removed from the motion code: persistent appearance is delegated to the reference frame, while the remaining temporal variation is compressed jointly. This separation is valuable, but it leaves the internal organization of motion capacity unresolved. Broad scene evolution and fine-grained, frame-specific structures still share one temporal sampling pattern and token geometry, despite requiring different temporal support.

This one-size-fits-all geometry is inefficient under a compact bottleneck. Dense temporal support repeatedly encodes predictable evolution, whereas temporal pooling can blur or discard fine-grained structures whose exact configuration varies across frames. The key question is therefore not only how much motion capacity to retain, but \emph{how a fixed budget should balance broad evolution with frame-specific detail}.

\begin{figure}[t]
  \centering
  \includegraphics[width=\columnwidth]{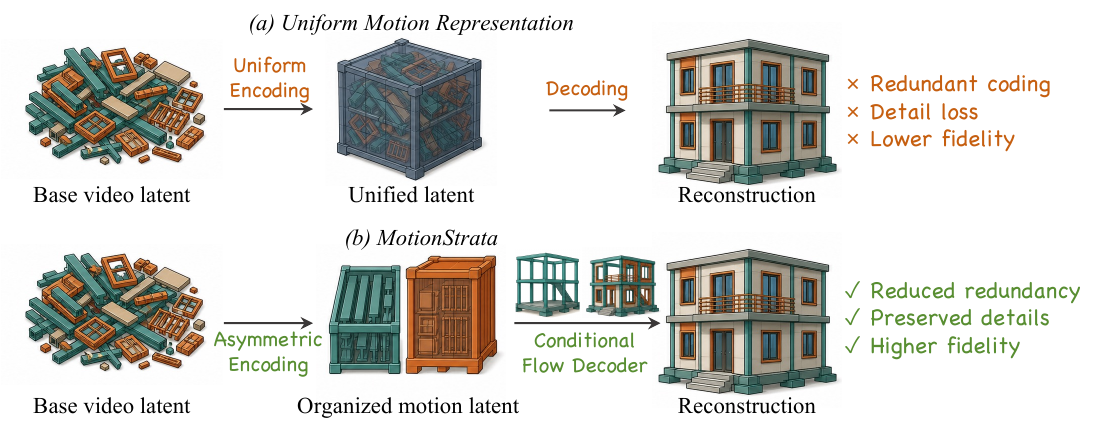}
  \caption{MotionStrata divides a fixed motion budget into compressed Global and frame-aligned Detailed groups.}
  \label{fig:concept}
\end{figure}
Our observation is that broad evolution can be summarized on a shorter temporal grid, while frame-local  benefit from dense support. Based on this observation, MotionStrata divides the same motion dimensionality into temporally compressed Global Motion and frame-aligned Detailed Motion. Complementary 3D frequency routing provides low-frequency and residual views to the two encoders, and a Global-first curriculum learns the compact level before introducing the frame-aligned level. At inference, both motion codes jointly condition a shared flow-matching decoder with the first-frame content anchor. The hierarchy is therefore defined by asymmetric temporal support and ordered learning, rather than by a sequential decoder or semantic disentanglement.
Guided by this observation, we introduce MotionStrata, which represents motion as a progression from broad evolution to frame-local refinement. Global Motion forms a compact temporal scaffold; Detailed Motion complements it with changes that require denser support. We realize this ordering through latent routing and coarse-to-fine learning, then reconstruct the video with both motion groups and the first-frame content anchor. The goal is not to force motion into independent semantics, but to preserve heterogeneous temporal patterns more effectively within the same compact interface. We evaluate this principle with motion dimensionality, content conditioning, decoder, sampler, and optimization budget controlled. The resulting comparisons isolate motion organization from capacity and decoder strength. Across reconstruction, controlled ablations, unseen domains, longer and higher-resolution videos, and downstream generation, the evidence consistently favors a hierarchical motion representation over a homogeneous one. Our main contributions are summarized as follows:
\begin{itemize}
    \item \textbf{A representation perspective.} We identify the internal organization of motion capacity as an overlooked bottleneck in compact video autoencoding: temporal patterns with different degrees of redundancy should not be forced into a homogeneous latent geometry.
    \item \textbf{A hierarchical motion latent.} We introduce MotionStrata, which organizes a fixed motion budget as a Global scaffold followed by Detailed refinement, preserving broad evolution and brief changes without increasing the motion interface.
    \item \textbf{Evidence under controlled budgets.} Matched comparisons establish the benefit of hierarchical organization, while broader evaluations examine operating range.
\end{itemize}

\section{Related Work}

\textbf{Video autoencoding and content--motion factorization.} Video VAEs add temporal convolution or attention to image autoencoders, producing compact latents for video generation~\cite{rombach2022high,lin2024open,agarwal2025cosmos,zhao2024cv,ge2022tats,yu2023magvit,yu2024magvitv2,wang2024omnitokenizer,sun2024uniavatar,yin2026holo}. Content--motion methods exploit a stronger prior: persistent appearance can be anchored by a reference frame while a smaller code describes temporal change~\cite{villegas2017decomposing,tulyakov2018mocogan}. Reducio~\cite{tian2024reducio} follows this setting, and VidTwin~\cite{wang2024vidtwin} separates structure from dynamics. MotionStrata operates inside the motion code left by this factorization. It organizes a fixed motion budget as a Global-to-Detailed hierarchy with different temporal support.

\noindent \textbf{Hierarchical representation and token allocation.} Classical Laplacian pyramids organize images by spatial frequency, while pyramid flow networks estimate displacement through coarse-to-fine refinement~\cite{burt1983laplacian,sun2018pwcnet,razavi2019vqvae2,vahdat2020nvae,hu2026preserve}. MotionStrata brings this hierarchical principle into a learned video bottleneck without decoding an explicit image pyramid. Frequency-derived views route evidence to Global and Detailed encoders; asymmetric temporal support and ordered training establish the hierarchy under a shared budget. A separate line varies the amount of representation allocated to each sample or time step: ElasticTok~\cite{yanelastictok2024}, AdapTok~\cite{liadaptok2025}, DLFR-VAE~\cite{yuandlfr2025}, InfoTok~\cite{yeinfotok2025}, One-DVA~\cite{tengonedva2026}, and EVATok~\cite{xiongevatok2026} learn adaptive frame rates, information-aware allocation, or variable token lengths. These methods ask \emph{how much} capacity to allocate; MotionStrata holds total motion dimensionality fixed and asks how that capacity should be organized.

\noindent \textbf{Generative decoding under compact bottlenecks.} Conditional diffusion and rectified-flow models can reconstruct details that remain ambiguous under compressed conditioning~\cite{peebles2023scalable,liu2022flow,preechakul2022diffusion,hudson2024soda}. FlowMo~\cite{sargent2025flow} studies perceptual image reconstruction with a Transformer diffusion autoencoder; we include it as an image-only efficiency reference. MotionStrata instead decodes video latents from the first frame and both motion groups. Our central controls share the decoder, sampler, and update budget to isolate the representation.

\section{Method}
\label{sec:method}

\begin{figure*}[t]
  \centering
  \includegraphics[width=0.92\textwidth]{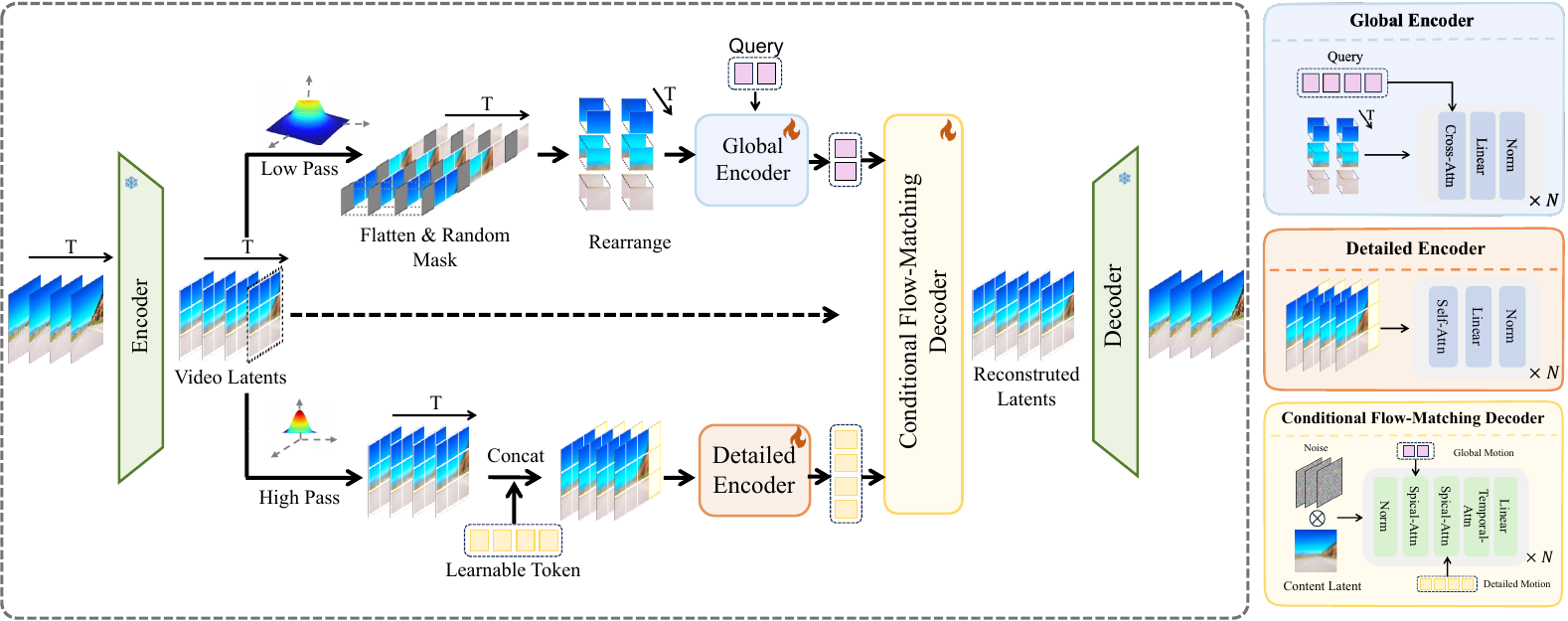}
  \caption{MotionStrata organizes a fixed motion budget as a Global-to-Detailed hierarchy. The solid arrows provide Global Motion, Detailed Motion, and first-frame content conditioning. The dashed path carries the flow state $z_t$ during training and starts from Gaussian noise at inference; the decoder never receives the clean target latent as conditioning.}
  \label{fig:pipeline}
\end{figure*}

MotionStrata follows one principle: compact motion should be represented as ordered abstraction and refinement. Figure~\ref{fig:pipeline} implements this in three steps: routed views provide evidence to the two levels, asymmetric query encoders establish Global Motion and Detailed Motion under a fixed budget, and a conditional flow model decodes them jointly.

\subsection{Representation Interface}

Given a video $x\in\mathbb{R}^{3\times F\times H\times W}$, we normalize each frame to $[-1,1]$ and encode it independently with the frozen SD-VAE AutoencoderKL~\cite{rombach2022high}. We draw once from its posterior and apply the Stable Diffusion latent scaling:
\begin{equation}
\begin{aligned}
 \tilde z_i&\sim q_{\mathcal{E}}(\tilde z_i\mid x_i),\quad
 z_i=0.18215\,\tilde z_i,\\
 z&=(z_0,\ldots,z_{f-1})\in\mathbb{R}^{f\times h\times w\times c}.
\end{aligned}
 \label{eq:base_latent}
\end{equation}
The learned motion contains Global Motion $u_g\in\mathbb{R}^{f_g\times n_g\times c_g}$ and Detailed Motion $u_d\in\mathbb{R}^{f\times n_d\times c_d}$. The first-frame latent $c_0=z_0\in\mathbb{R}^{h\times w\times c}$ provides appearance and layout evidence but has no temporal receptive field. After reconstructing $\hat z$, the frozen frame-wise decoder $\mathcal{G}$ receives $\hat z/0.18215$ and produces $\hat x$ (Details are given in Supp. Sec.~2).

\subsection{Hierarchical Motion Encoding}

Global Motion establishes a temporally compact scaffold, and Detailed Motion retains frame alignment to refine that scaffold. We apply a three-dimensional FFT over $(f,h,w)$ for each channel. A binary centered mask $P_k$ retains a low-frequency cuboid whose cutoff is normalized by the Nyquist frequency of each axis. Using the default $k=1/4$, we define
\begin{equation}
 z_{\mathrm{low}}=\mathcal{F}^{-1}(\mathcal{F}(z)\odot P_k),\qquad
 z_{\mathrm{high}}=z-z_{\mathrm{low}}.
 \label{eq:frequency_split}
\end{equation}

\noindent\textbf{Global Motion.} The Global level summarizes broad evolution on a shorter temporal query grid. Its routed input $z_{\mathrm{low}}$ emphasizes evidence that changes gradually across neighboring frames. Let $\tilde z_{\mathrm{low}}=\operatorname{Flatten}_{h,w}(z_{\mathrm{low}})\in\mathbb{R}^{f\times hw\times c}$. During training, a random spatial mask $M$ removes 30\% of spatial tokens to discourage appearance copying. Learnable queries $q_g\in\mathbb{R}^{f_g\times n_g\times c_g}$ then aggregate the visible features:
\begin{equation}
 r_g=E_g(M\odot\tilde z_{\mathrm{low}},q_g)
 \in\mathbb{R}^{f_g\times n_g\times c_g},\qquad f_g=f/2.
 \label{eq:global_encoder}
\end{equation}
Each block in $E_g$ uses $q_g$ as queries and the visible latent tokens as keys and values. The shorter grid avoids repeatedly encoding similar states and biases the fixed Global budget toward evolution shared across neighboring frames.

\noindent\textbf{Detailed Motion.} The Detailed level refines this scaffold with evidence that can vanish under temporal pooling. We therefore retain one query set per latent frame, $q_d\in\mathbb{R}^{f\times n_d\times c_d}$. For frame $i$, an MLP $\pi_d$ projects the flattened residual features from $c$ to $c_d$ channels. The projected features and queries are concatenated along the token axis:
\begin{align}
 p_d^i &= \pi_d(\operatorname{Flatten}_{h,w}(z_{\mathrm{high}}^i))
 \in\mathbb{R}^{hw\times c_d},\\
 h_d^i &= \operatorname{Concat}_{\mathrm{token}}[q_d^i,p_d^i],\\
 r_d^i &= \operatorname{QueryOut}(E_d(h_d^i))
 \in\mathbb{R}^{n_d\times c_d}.
 \label{eq:detailed}
\end{align}
Here $E_d$ is a self-attention stack and $\operatorname{QueryOut}$ retains only the outputs at the $n_d$ query positions. Stacking $\{r_d^i\}_{i=1}^{f}$ gives $r_d\in\mathbb{R}^{f\times n_d\times c_d}$. Unlike the Global grid, this branch preserves latent-frame alignment and refines local dynamics within the broad evolution carried by Global Motion.

\noindent\textbf{Motion posterior.} Each encoder parameterizes a diagonal Gaussian posterior through learned mean and variance:
\begin{align}
 (\mu_g,\log\sigma_g^2)&=H_g(r_g),&
 u_g&=\mu_g+\sigma_g\odot\epsilon_g,\nonumber\\
 (\mu_d,\log\sigma_d^2)&=H_d(r_d),&
 u_d&=\mu_d+\sigma_d\odot\epsilon_d,
 \label{eq:motion_posterior}
\end{align}
where $\epsilon_g,\epsilon_d\sim\mathcal{N}(0,I)$. The sampled $u_g$ and $u_d$ form the transmitted motion interface.
The fixed $k=1/4$ split defines the model. We test neighboring fixed cutoffs without changing the number or dimensionality of motion tokens.

\subsection{Conditional Flow-Matching Decoder}
Under a compact bottleneck, $(u_g,u_d,c_0)$ may not uniquely specify every base-latent detail. A deterministic regressor can average over these possibilities and oversmooth difficult regions. We instead model the conditional reconstruction distribution with a DiT-based flow decoder~\cite{peebles2023scalable,liu2022flow}. For $t\sim\mathcal{U}[0,1]$ and $\epsilon\sim\mathcal{N}(0,I)$, the noise-to-data path is
\begin{equation}
 z_t=(1-t)\epsilon+tz,\quad v=z-\epsilon,\quad
 \hat v=D_\theta(z_t,t;u_g,u_d,c_0).
 \label{eq:flow}
\end{equation}
Thus, $t=0$ is Gaussian noise and $t=1$ is the clean base latent. The clean target $z$ defines the training path but is never supplied as decoder conditioning.
The decoder first patchifies $z_t$ and the temporally repeated content anchor:
\begin{equation}
 a_t=\operatorname{Concat}_{\mathrm{token}}
 [\operatorname{Patch}(z_t),\operatorname{Patch}(\operatorname{Repeat}_f(c_0))].
 \label{eq:decoder_input}
\end{equation}
Projected Global and Detailed tokens interact with the reconstruction state in every decoder block. Spatial attention models within-frame structure, while temporal attention aligns corresponding positions across frames. A linear head predicts velocity only for the $z_t$ state tokens, which are unpatchified to obtain $\hat v$. This makes $c_0$, $u_g$, and $u_d$ explicit conditions rather than additional clean-video inputs. Full architectural and solver details are provided in Appendix~S2.

\subsection{Coarse-to-Fine Training}
\begin{table*}[t]
\centering
\small
\setlength{\tabcolsep}{2.8pt}
\begin{tabular}{llccccccc}
\toprule
Family & Method & $r_{\mathrm{motion}}\downarrow$ & $r_{\mathrm{total}}\downarrow$ & PSNR$\uparrow$ & SSIM$\uparrow$ & LPIPS$\downarrow$ & rFID$\downarrow$ & rFVD$\downarrow$ \\
\midrule
\multirow{5}{*}{Content}
 & SD2.1-VAE~\cite{rombach2022high} & -- & 2.08\% & 26.684 & 0.813 & 0.106 & 5.591 & 81.785 \\
 & WF-VAE~\cite{li2025wfvae} & -- & 1.04\% & \rankFirst{29.374} & \rankThird{0.827} & 0.103 & \rankThird{4.137} & 78.883 \\
 & Cosmos-VAE~\cite{agarwal2025cosmos} & -- & 2.08\% & 28.481 & \rankSecond{0.830} & 0.105 & 6.449 & 121.986 \\
 & CV-VAE~\cite{zhao2024cv} & -- & 0.53\% & \rankThird{29.008} & 0.825 & \rankSecond{0.099} & \rankSecond{4.030} & 86.162 \\
 & LeanVAE~\cite{cheng2025leanvae} & -- & 2.08\% & 28.070 & 0.819 & 0.106 & 5.367 & 110.379 \\
\midrule
\multirow{7}{*}{Content--motion}
 & iVideoGPT~\cite{wu2024ivideogpt} & 0.46\% & 1.50\% & 20.713 & 0.559 & 0.326 & 18.677 & 356.904 \\
 & CMD~\cite{yu2024efficient} & \rankSecond{0.13\%} & \rankSecond{0.26\%} & 26.553 & 0.795 & 0.110 & 11.664 & 98.623 \\
 & Reducio~\cite{tian2024reducio} & \rankSecond{0.13\%} & 6.38\% & 26.484 & 0.786 & 0.112 & 7.334 & 108.936 \\
 & VidTwin~\cite{wang2024vidtwin} & 0.20\% & \rankFirst{0.20\%} & 25.530 & 0.799 & 0.119 & 11.941 & 131.059 \\
\cmidrule(l){2-9}
 & MotionStrata (S) & \rankFirst{0.07\%} & \rankFirst{0.20\%} & 28.365 & 0.808 & 0.109 & 5.425 & \rankThird{74.005} \\
 & MotionStrata & \rankThird{0.15\%} & \rankThird{0.28\%} & 28.861 & 0.826 & \rankThird{0.101} & 5.227 & \rankSecond{71.278} \\
 & MotionStrata (L) & 0.28\% & 0.41\% & \rankSecond{29.347} & \rankFirst{0.834} & \rankFirst{0.096} & \rankFirst{4.002} & \rankFirst{61.941} \\
\bottomrule
\end{tabular}
\caption{Reconstruction on 16-frame, $256\!\times\!256$ WebVid clips. $r_{\mathrm{motion}}$ counts tensors designed to carry temporal variation; $r_{\mathrm{total}}$ counts every reconstruction input. Both divide scalar dimensionality by $3FHW$ and are not bitrates. Content VAEs do not expose a separately identifiable motion code. Best, second-best, and third-best values are shown in bold, underlined, and italics.}
\label{tab:main_results}
\end{table*}

The conditional flow objective is
\begin{equation}
 \mathcal{L}_{\mathrm{flow}}
 =\mathbb{E}_{z,t,\epsilon}
 \left[\lVert v-\hat v\rVert_2^2\right].
 \label{eq:flow_loss}
\end{equation}
\begin{equation}
\begin{aligned}
 \mathcal{L}_{\mathrm{KL}}=\lambda_{\mathrm{KL}}\big[&
 \mathrm{KL}\!\left(q_g(u_g\mid r_g)\,\|\,\mathcal{N}(0,I)\right)\\
 &+\mathrm{KL}\!\left(q_d(u_d\mid r_d)\,\|\,\mathcal{N}(0,I)\right)\big],
\end{aligned}
\label{eq:motion_kl}
\end{equation}
and the total objective, with KL terms applied to the active motion posteriors at each stage, is
\begin{equation}
 \mathcal{L}=\mathcal{L}_{\mathrm{flow}}+\mathcal{L}_{\mathrm{KL}}.
 \label{eq:total_loss}
\end{equation}
Training follows the same hierarchy as the representation. Stage 1 trains the Global encoder and shared decoder for 20k updates using $z_{\mathrm{low}}$ as the flow target and $(u_g,c_0)$ as conditions. This establishes broad evolution before the decoder must resolve frame-local changes. Stage 2 freezes the Global encoder, introduces the Detailed encoder, and updates the Detailed encoder and shared decoder for 80k further steps against the full target $z$. Freezing preserves the Global scaffold while Detailed Motion learns frame-aligned refinements under the full-target objective. At inference, we draw one sample from each motion posterior and integrate the learned velocity ODE for $S$ steps to obtain $\hat z$.

\section{Experiments}
\label{sec:experiments}
\subsection{Experimental Setup}
\textbf{Datasets.} We train MotionStrata on WebVid-10M~\cite{bain2021frozen} and evaluate in-domain reconstruction on its validation set. For zero-shot reconstruction, we apply the frozen checkpoint to UCF-101~\cite{soomro2012ucf101} and RealEstate10K~\cite{zhou2018stereo} without finetuning. The stress tests additionally use 500 long-eligible WebVid clips and 500 native-resolution RealEstate10K clips. Downstream generation is evaluated on UCF-101 and SkyTimelapse~\cite{8578349} (protocols in Appendix~S1).
\begin{figure}[t]
  \centering
  \includegraphics[width=\columnwidth]{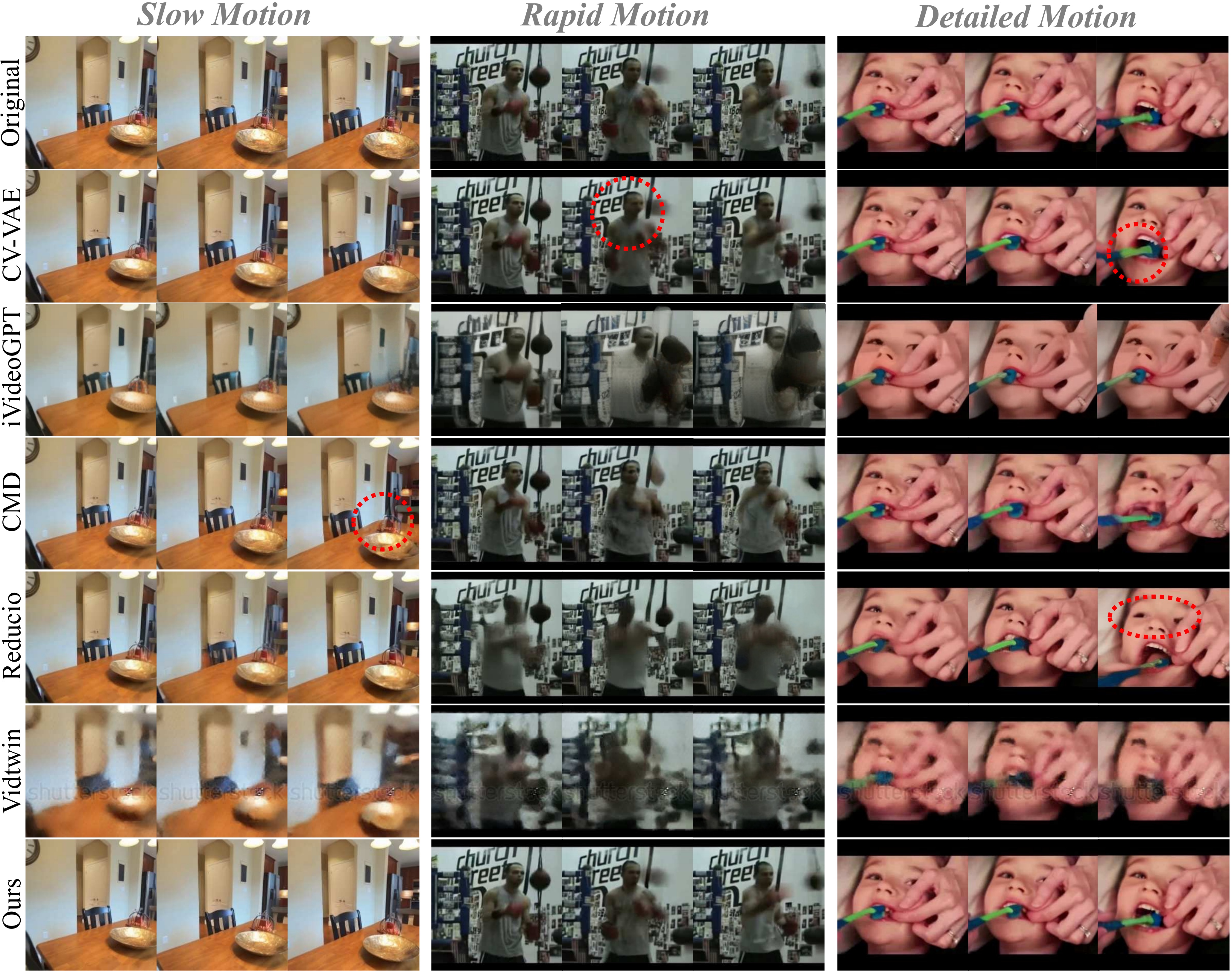}
  \caption{Qualitative reconstruction comparison across diverse temporal changes. MotionStrata better preserves local structure under the evaluated compact motion budgets.}
  \label{fig:reconstruction}
\end{figure}

\noindent\textbf{Training and inference.} Unless stated otherwise, all experiments use 16-frame, $256\!\times\!256$ clips at 8 fps; the resolution test handles native $832\!\times\!480$ inputs with overlapping $256^2$ tiles. The frame-wise base VAE produces a first-frame anchor $c_0\in\mathbb{R}^{1\times32\times32\times4}$. In temporal order, $u_g\in\mathbb{R}^{8\times8\times8}$ and $u_d\in\mathbb{R}^{16\times8\times32}$, totaling 4,608 motion scalars. Unless explicitly marked (S) or (L), ``MotionStrata'' refers to this configuration. The (S) variant uses
$u_g\in\mathbb{R}^{4\times8\times8}$ and
$u_d\in\mathbb{R}^{16\times8\times16}$, totaling 2,304 motion scalars,
whereas (L) uses
$u_g\in\mathbb{R}^{8\times8\times8}$ and
$u_d\in\mathbb{R}^{16\times16\times32}$, totaling 8,704 motion scalars. Each motion encoder has eight Transformer layers and eight heads; the decoder has 16 blocks. We train on eight A800 GPUs with Adam at $10^{-4}$ and use 20-step Euler sampling by default (full details in Appendices~S1--S2).

\noindent\textbf{Evaluation metrics and protocol.} We measure reconstruction distortion with PSNR~\cite{hore2010image} and SSIM~\cite{wang2004image}, perceptual fidelity with LPIPS~\cite{zhang2018unreasonable} and rFID~\cite{heusel2017gans}, and temporal fidelity with rFVD~\cite{unterthiner2018towards}. Downstream generation reports FID and FVD; diagnostic metrics are detailed in Appendices~S1 and~S3.

\noindent\textbf{Representation accounting.} To distinguish the learned motion code from all decoder conditions, we report
\begin{equation}
 r_{\mathrm{motion}}=\frac{|u_g|+|u_d|}{3FHW},\qquad
 r_{\mathrm{total}}=\frac{|c_0|+|u_g|+|u_d|}{3FHW}.
 \label{eq:ratios}
\end{equation}
The first counts the two motions; the second also includes the content anchor. For MotionStrata, $r_{\mathrm{motion}}=0.146\%$ and $r_{\mathrm{total}}=0.277\%$ (full accounting in Appendix~S2).

\noindent\textbf{Comparisons and controls.} We compare with two families of video autoencoders. The content--motion family includes iVideoGPT~\cite{wu2024ivideogpt}, CMD~\cite{yu2024efficient}, Reducio~\cite{tian2024reducio}, and VidTwin~\cite{wang2024vidtwin}, which we retrain under the same clip protocol. The content-VAE family includes SD2.1-VAE~\cite{rombach2022high}, WF-VAE~\cite{li2025wfvae}, Cosmos-VAE~\cite{agarwal2025cosmos}, CV-VAE~\cite{zhao2024cv}, and LeanVAE~\cite{cheng2025leanvae}. We use the former as the primary comparison under a similar setting and the latter as external reference points on the broader quality--compactness landscape.

\begin{table}[t]
\centering
\footnotesize
\setlength{\tabcolsep}{2.0pt}
\resizebox{\columnwidth}{!}{%
\begin{tabular}{llcccc}
\toprule
Representation control & Optimization & PSNR$\uparrow$ & SSIM$\uparrow$ & rFID$\downarrow$ & rFVD$\downarrow$ \\
\midrule
Flat latent & Single-stage & 25.791 & 0.762 & 7.135 & 89.197 \\
Random split & Single-stage & 25.807 & 0.775 & 7.257 & 86.175 \\
Learned split & Single-stage & \rankThird{27.845} & 0.792 & \rankThird{5.529} & \rankThird{78.240} \\
\midrule
Full-band + asym. grids & Two-stage, G frozen & 27.422 & 0.796 & 5.641 & 81.316 \\
3D FFT + sym. grids & Two-stage, G frozen & 27.583 & \rankThird{0.803} & \rankSecond{5.514} & 79.812 \\
Temporal FFT + asym. grids & Two-stage, G frozen & \rankSecond{28.314} & \rankSecond{0.813} & 5.563 & \rankSecond{75.914} \\
Spatial FFT + asym. grids & Two-stage, G frozen & 27.126 & 0.791 & 6.118 & 83.127 \\
3D FFT + asym. grids (ours) & Two-stage, G frozen & \rankFirst{28.861} & \rankFirst{0.826} & \rankFirst{5.227} & \rankFirst{71.278} \\
\bottomrule
\end{tabular}}
\caption{Ablations use the same content code, motion dimensionality, decoder, sampler, encoder depth, and 100k total updates as MotionStrata. The upper block gives single-stage organization baselines. The lower block fixes two-stage optimization with Global frozen in Stage 2, isolating routing support and query-grid geometry.}
\label{tab:matched_ablation}
\end{table}

\begin{table}[t]
\centering
\small
\setlength{\tabcolsep}{2.2pt}
\begin{tabular}{lcccc}
\toprule
Training strategy & PSNR$\uparrow$ & SSIM$\uparrow$ & rFID$\downarrow$ & rFVD$\downarrow$ \\
\midrule
Single-stage (100k) & \rankThird{27.603} & \rankThird{0.791} & \rankThird{6.827} & \rankThird{92.831} \\
Two-stage, G trainable & \rankSecond{28.527} & \rankSecond{0.815} & \rankSecond{5.398} & \rankSecond{75.921} \\
Two-stage, G frozen & \rankFirst{28.861} & \rankFirst{0.826} & \rankFirst{5.227} & \rankFirst{71.278} \\
\bottomrule
\end{tabular}
\caption{Ablation for the complete 3D FFT + asymmetric-grid architecture. All rows use 100k updates; two-stage rows allocate 20k updates to Stage 1 and 80k to Stage 2.}
\label{tab:training_schedule}
\end{table}

\subsection{Quantitative \& Qualitative Comparison}

\noindent\textbf{Quantitative results.} Across three motion budgets, MotionStrata improves quality as the structured motion capacity increases (Table~\ref{tab:main_results}). The small, default, and large variants use motion ratios of 0.07\%, 0.15\%, and 0.28\%, and total ratios of 0.20\%, 0.28\%, and 0.41\%, respectively. The default reaches 28.861 PSNR and 71.278 rFVD, improving distortion, perceptual quality, and temporal fidelity over the evaluated content--motion baselines. The large model further reaches 29.347 PSNR and 61.941 rFVD.

\noindent\textbf{Qualitative results.} Figure~\ref{fig:reconstruction} complements the main table with clips containing diverse temporal changes. MotionStrata better preserves local structures and transient details than other content and content--motion VAEs while maintaining coherent scene evolution (more results in Appendix~S4).

\subsection{Ablation Studies}
\label{sec:ablation}

\noindent\textbf{Motion organization.}
The upper block of Table~\ref{tab:matched_ablation} compares one homogeneous stream (Flat), a fixed random two-branch assignment (Random), and a learned two-way router (Learned). All controls match dimensionality, architecture, sampler, and updates. Only the learned partition gives a clear gain, showing that branch count alone is insufficient (full definitions in Appendix~S2).
\begin{table}[t]
\centering
\small
\setlength{\tabcolsep}{3.1pt}
\begin{tabular}{lcccc}
\toprule
Split & PSNR$\uparrow$ & SSIM$\uparrow$ & rFID$\downarrow$ & rFVD$\downarrow$ \\
\midrule
Flat, no split & 25.791 & 0.762 & 7.135 & 89.197 \\
Fixed $k=1/8$ & \rankThird{28.112} & \rankFirst{0.827} & \rankSecond{5.268} & \rankThird{73.936} \\
Fixed $k=1/4$ & \rankFirst{28.861} & \rankSecond{0.826} & \rankFirst{5.227} & \rankFirst{71.278} \\
Fixed $k=3/8$ & \rankSecond{28.554} & \rankThird{0.811} & \rankThird{5.371} & \rankSecond{73.842} \\
\bottomrule
\end{tabular}
\caption{Routing-cutoff sensitivity at a fixed motion budget.}
\label{tab:cutoff_analysis}
\end{table}

\begin{table}[t]
\centering
\small
\setlength{\tabcolsep}{3.2pt}
\begin{tabular}{lccc}
\toprule
Reconstruction condition & PSNR$\uparrow$ & LPIPS$\downarrow$ & rFVD$\downarrow$ \\
\midrule
Content anchor only
& 19.630
& 0.231
& 194.870 \\
Anchor + Global only
& \rankThird{23.913}
& \rankThird{0.175}
& \rankThird{126.530} \\
Anchor + Detailed only
& \rankSecond{26.572}
& \rankSecond{0.128}
& \rankSecond{94.310} \\
Anchor + Global + Detailed
& \rankFirst{28.861}
& \rankFirst{0.101}
& \rankFirst{71.278} \\
\bottomrule
\end{tabular}
\caption{Stream-removal sensitivity of Global and Detailed Motion. Single-stream rows zero the omitted condition.}
\label{tab:pathway_ablation}
\end{table}

\begin{table}[t]
\centering
\footnotesize
\setlength{\tabcolsep}{2.6pt}
\resizebox{\columnwidth}{!}{%
\begin{tabular}{lcccc}
\toprule
Condition & Low-band$\downarrow$ & High-band$\downarrow$ & Homography-fit EPE$\downarrow$ & Residual EPE$\downarrow$ \\
\midrule
Both streams & \rankFirst{0.119} & \rankFirst{0.567} & \rankFirst{1.187} & \rankFirst{1.130} \\
$u_g=0$ & \rankThird{0.866} & \rankSecond{0.832} & \rankThird{2.070} & \rankSecond{1.249} \\
$u_d=0$ & \rankSecond{0.384} & \rankThird{0.978} & \rankSecond{1.701} & \rankThird{1.391} \\
\bottomrule
\end{tabular}}
\caption{Post-encoding stream removal on a fixed 500-clip manifest. Band values are base-latent NRMSE; RAFT flow EPE is decomposed by fitted homography. Relative degradation measures the asymmetric contribution of each level.}
\label{tab:band_zeroout}
\end{table}

\noindent\textbf{Hierarchy components.}
The lower block fixes two-stage training and varies structure. Full-band removes routing, symmetric grids make both query grids frame-aligned, and one-axis FFT splits only the named axes. Complete 3D routing with asymmetric grids performs best.

\begin{figure*}[t]
  \centering
  \includegraphics[width=0.95\textwidth]{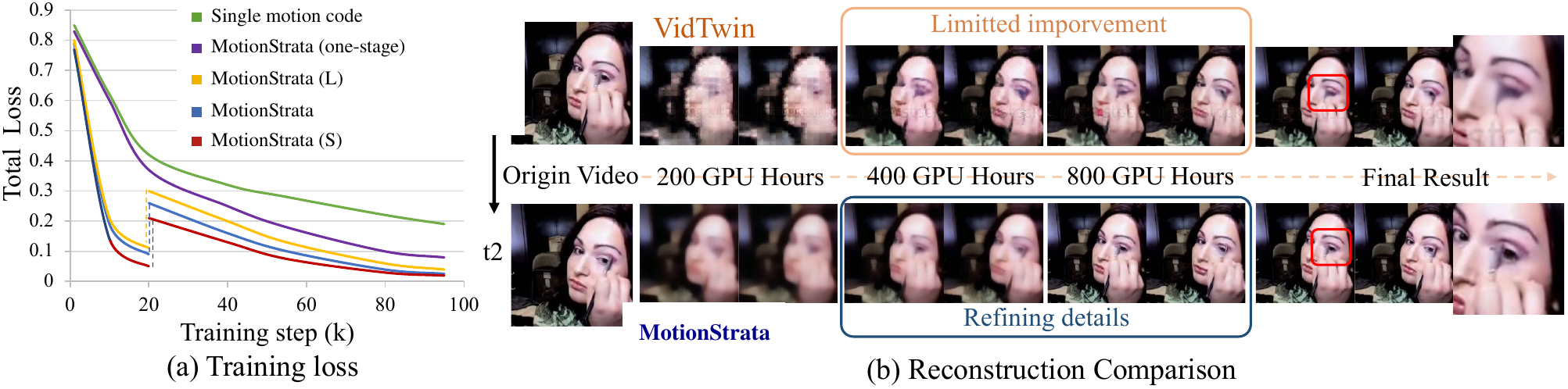}
  \caption{Optimization behavior. Left: reconstruction loss for a flat motion code, single-stage MotionStrata, and three MotionStrata budgets. Right: intermediate reconstructions illustrate how hierarchical training progressively restores fine details. The curves diagnose optimization dynamics.}
  \label{fig:training_dynamics}
\end{figure*}

\noindent\textbf{Training strategy.}
Table~\ref{tab:training_schedule} fixes the complete architecture and varies only its training schedule. Two-stage training improves substantially over single-stage training, and freezing Global Motion in Stage 2 gives the strongest temporal fidelity. This is consistent with preserving the coarse scaffold while Detailed Motion progressively refines the reconstruction. Figure~\ref{fig:training_dynamics} complements the final metrics with training curves and intermediate reconstructions.

\noindent\textbf{Cutoff sensitivity.}
Table~\ref{tab:cutoff_analysis} varies only the cutoff under a fixed motion budget and decoder. The default $k=1/4$ gives the best PSNR, rFID, and rFVD, while $k=1/8$ is marginally better in SSIM. Both neighboring cutoffs remain clearly stronger than the flat latent, showing that the benefit is not tied to one exact boundary.

\subsection{In-Depth Analysis}

\begin{figure}[t]
  \centering
  \includegraphics[width=0.95\columnwidth]{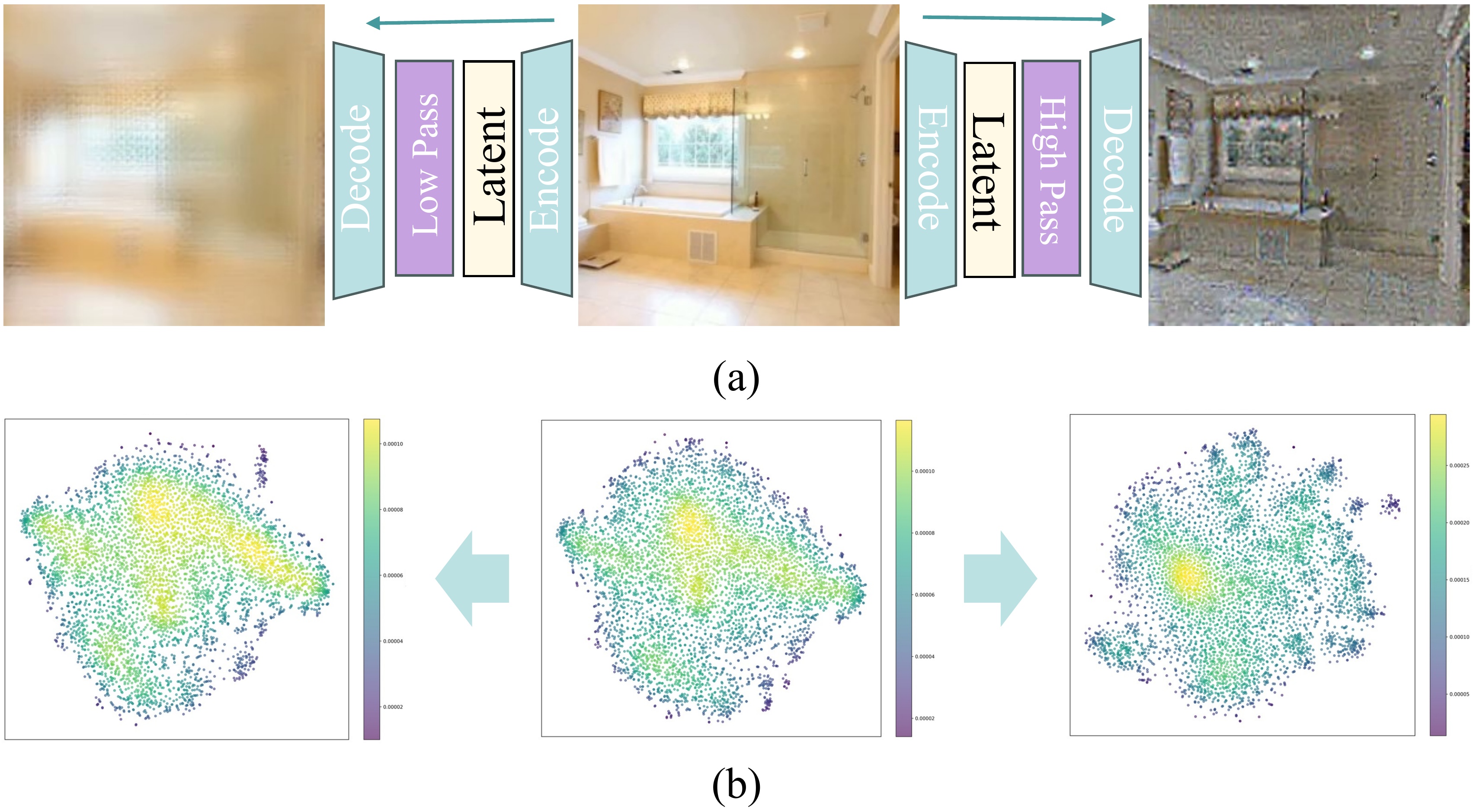}
  \vspace{-3mm}
  \caption{Frequency views in the latent space. The low-frequency component emphasizes smooth structure, while its residual accentuates higher-frequency evidence.}
  \label{fig:freq_vis}
\end{figure}

\noindent\textbf{Frequency routing.}
Figure~\ref{fig:freq_vis} visualizes the frequency-derived inputs before motion encoding. The low-frequency component emphasizes smooth structure, whereas the exact residual accentuates higher-frequency evidence (spectral analysis in Appendix~S3).

\begin{figure}[t]
  \centering
  \includegraphics[width=0.95\columnwidth]{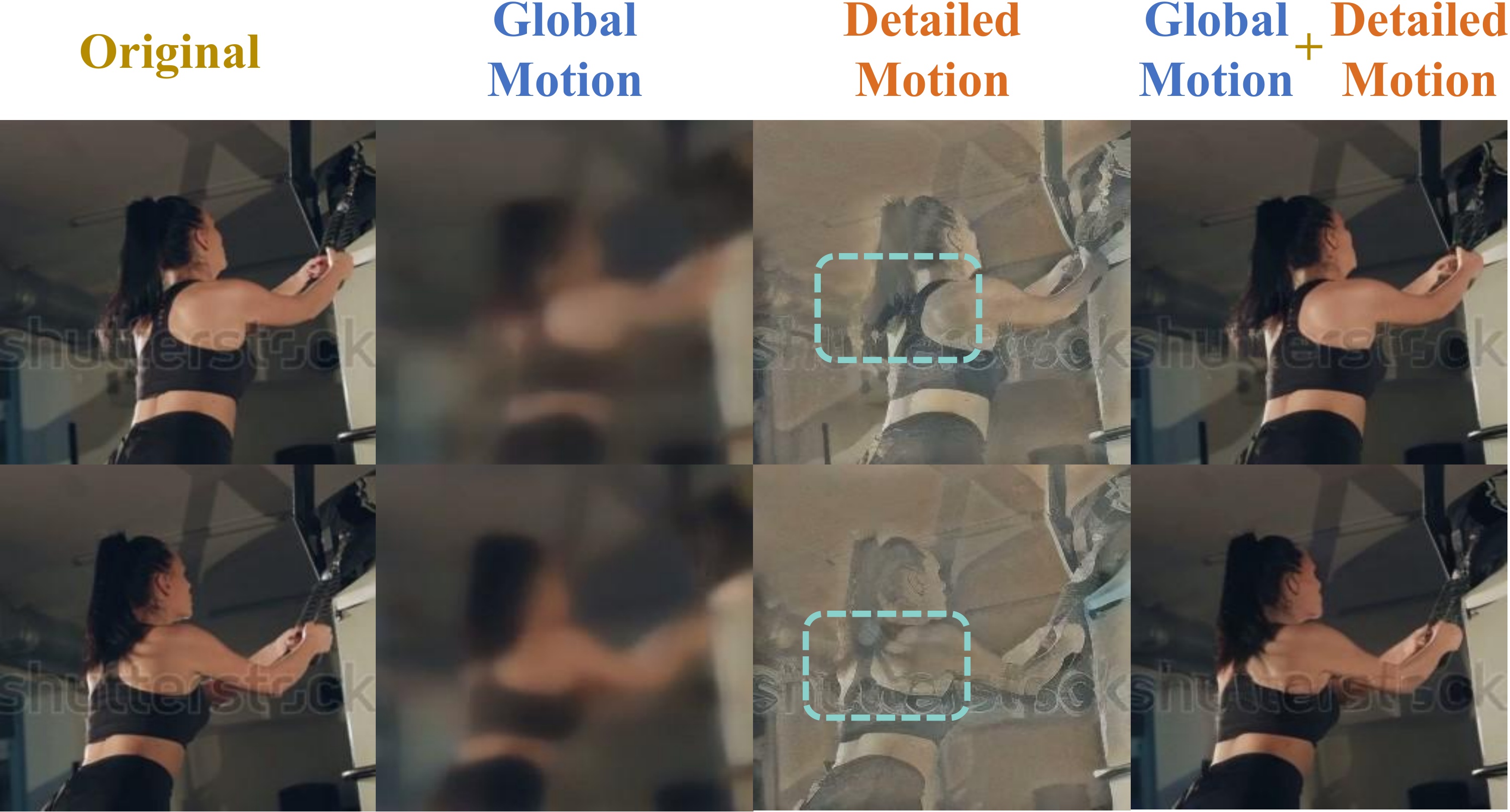}
  \caption{Single-stream decoding illustrates the hierarchy: Global Motion better preserves broad scene evolution, while Detailed Motion contributes more to frame-local structure.}
  \label{fig:latent_roles}
\end{figure}
\begin{figure}[t]
  \centering
  \includegraphics[width=0.95\columnwidth]{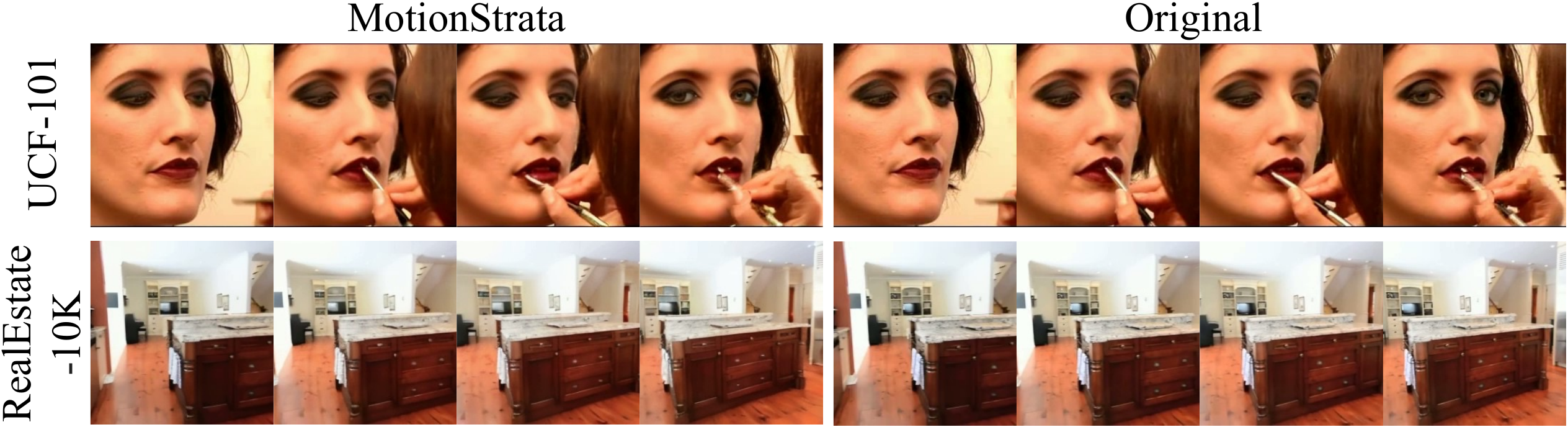}
  \vspace{-2mm}
  \caption{Zero-shot reconstruction on UCF-101 and RealEstate10K. WebVid-trained checkpoint is used without target-domain fine-tuning.}
  \label{fig:cross_dataset_reconstruction}
\end{figure}

\noindent\textbf{Group necessity.}
We zero each motion condition after encoding without retraining. Table~\ref{tab:pathway_ablation} shows that the content anchor alone is insufficient for accurate future reconstruction, each motion group improves reconstruction, and using both jointly performs best.

\noindent\textbf{Asymmetric motion contributions.}
Table~\ref{tab:band_zeroout} examines how each motion group contributes to reconstruction. Removing Global Motion causes increases in low-frequency latent error and flow error, whereas removing Detailed Motion more strongly affects high-frequency latent error and residual-flow error. Either removal degrades all metrics, confirming that the hierarchy benefits from both levels. Figure~\ref{fig:latent_roles} provides qualitative evidence (protocol in Appendix~S3).

\noindent\textbf{Evaluation beyond WebVid.}
We further evaluate the frozen WebVid-trained MotionStrata checkpoint on UCF-101 and RealEstate10K without fine-tuning. On fixed
2,000-clip manifests, it achieves 26.904 PSNR, 0.093 LPIPS, and 42.898 rFVD
on UCF-101, and 26.108 PSNR, 0.086 LPIPS, and 15.931 rFVD on RealEstate10K.
Figure~\ref{fig:cross_dataset_reconstruction} provides corresponding visual results across distinct appearance and motion distributions (details in Appendix~S1).



\noindent\textbf{Length and resolution.}
Table~\ref{tab:length_resolution} tests the default checkpoint beyond its 16-frame, $256^2$ training setting. It reports 16--64-frame reconstruction with segment-wise anchor refresh and native $832\!\times\!480$ tiled decoding; Figure~\ref{fig:length_resolution_qualitative} shows representative outputs (protocol in Appendix~S1).
With anchor refresh, the 64-frame setting remains close to the 16-frame setting in frame-level reconstruction quality.

\begin{table}[t]
\centering
\scriptsize
\setlength{\tabcolsep}{2.3pt}
\begin{tabular}{lrrr}
\toprule
Condition & PSNR$_{\rm VAE}\uparrow$ & LPIPS$_{\rm VAE}\downarrow$ & rFVD$_{\rm VAE}\downarrow$ \\
\midrule
16 frames & 28.189 & 0.105 & 65.395 \\
32 frames & 28.006 & 0.104 & 74.187 \\
64 frames & 27.100 & 0.107 & 56.795 \\
$832\!\times\!480$ tiled & 28.754 & 0.151 & 69.605 \\
\bottomrule
\end{tabular}
\caption{Length and resolution stress tests on fixed 500-clip manifests. Temporal rows use long-eligible WebVid clips with segment-wise anchor refresh; the native-resolution row uses RealEstate10K with overlapping tiled decoding. All metrics use frame-wise VAE reconstructions as reference.}
\label{tab:length_resolution}
\end{table}

\begin{figure}[t]
  \centering
  \includegraphics[width=\columnwidth]{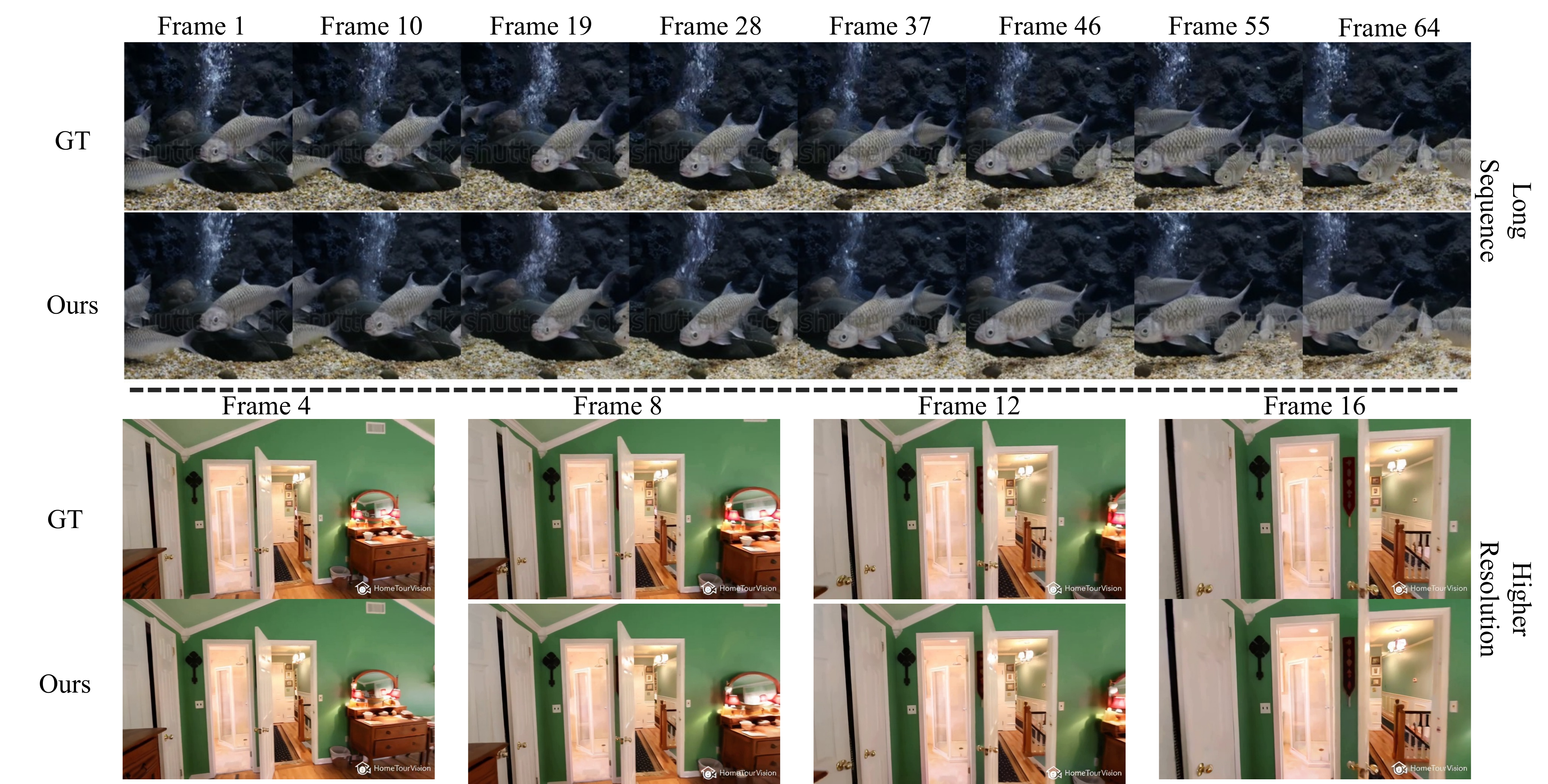}
  \caption{Stress-test reconstructions from the default checkpoint. Top: a 64-frame WebVid clip reconstructed with segment-wise anchor refresh. Bottom: overlapping-tile reconstruction of a native $832\!\times\!480$ RealEstate10K clip.}
  \label{fig:length_resolution_qualitative}
\end{figure}

\begin{figure}[t]
  \centering
  \includegraphics[width=\columnwidth]{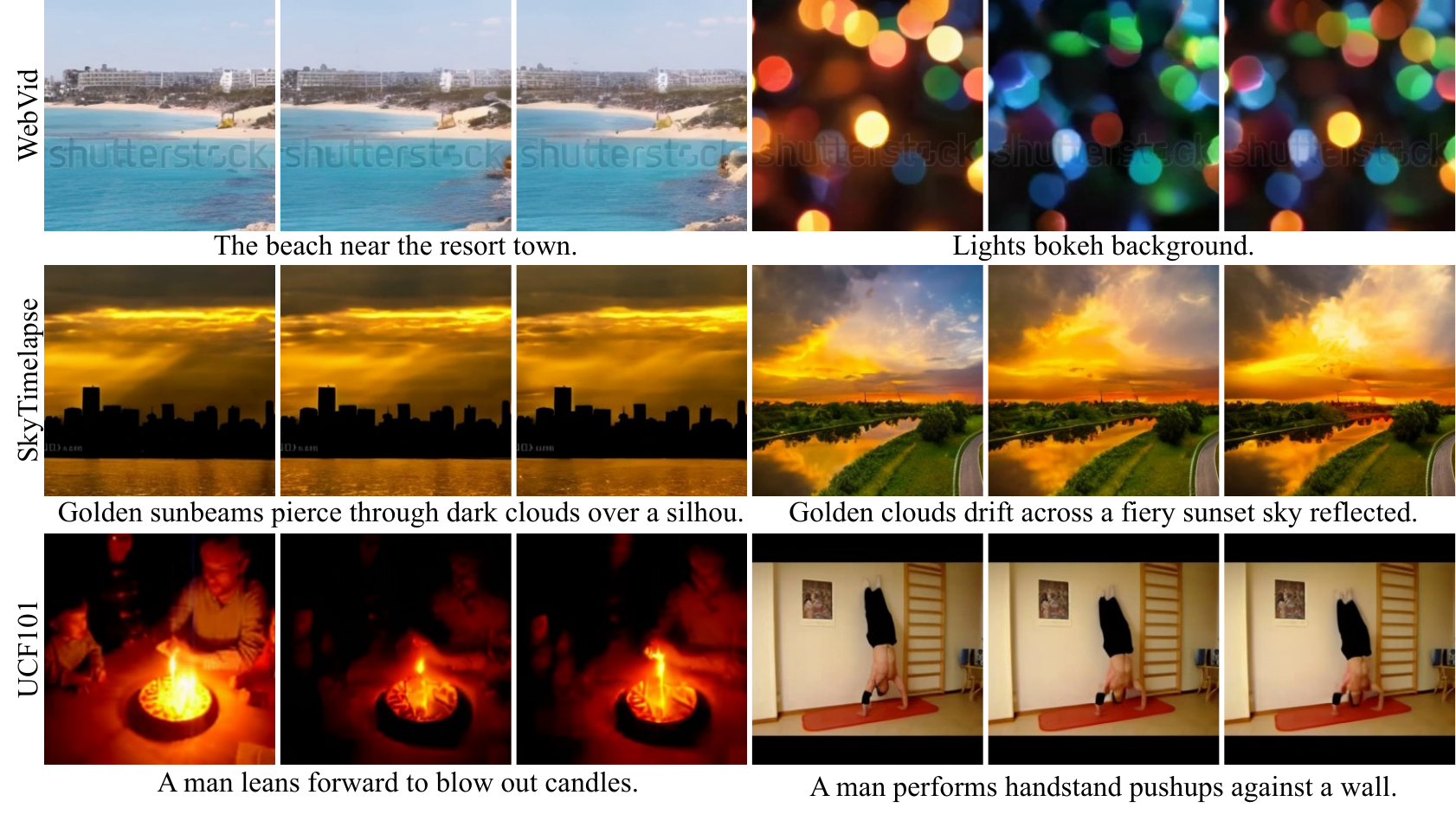}
  \caption{Representative MotionStrata generations on UCF-101 and SkyTimelapse under the downstream latent-generator protocol.}
  \label{fig:downstream_gen}
\end{figure}

\noindent\textbf{Generative compatibility.}
To verify that the learned motion latents are usable beyond reconstruction, we conduct a lightweight compatibility study in which a latent diffusion Transformer models the compressed motion-token space. Given the first-frame latent and a text or category condition, a Transformer-based conditional generator following the LatteMotion design~\cite{ma2024latte} predicts the motion tokens, which are then decoded by the frozen MotionStrata decoder. Table~\ref{tab:downstream} compares MotionStrata with published results from Video-LaViT, iVideoGPT, CMD, WF-VAE + Latte, and LeanVAE + Latte under their native settings. The results and examples in Figure~\ref{fig:downstream_gen} show that the compact hierarchy forms a learnable downstream interface; this experiment evaluates compatibility rather than state-of-the-art generation quality. More details are provided in Appendix~S5.

\begin{table}[t]
\centering
\small
\setlength{\tabcolsep}{2.0pt}
\resizebox{\columnwidth}{!}{%
\begin{tabular}{@{}lcccc@{}}
\toprule
 & \multicolumn{3}{c}{UCF-101} & SkyTimelapse \\
\cmidrule(lr){2-4}\cmidrule(l){5-5}
Model & FID$\downarrow$ & FVD$\downarrow$ & LPIPS$\downarrow$ & FVD$\downarrow$ \\
\midrule
Video-LaViT~\cite{jin2024video} & 25.822 & 204.661 & 0.133 & 154.637 \\
iVideoGPT~\cite{wu2024ivideogpt} & 28.304 & 254.881 & 0.162 & 196.329 \\
CMD~\cite{yu2024efficient} & 22.304 & 210.332 & \rankThird{0.125} & 130.374 \\
WF-VAE + Latte~\cite{li2025wfvae} & \rankThird{17.637} & \rankThird{192.334} & 0.128 & \rankThird{113.672} \\
LeanVAE + Latte~\cite{cheng2025leanvae} & \rankFirst{15.525} & \rankFirst{178.664} & \rankSecond{0.115} & \rankSecond{95.152} \\
MotionStrata & \rankSecond{16.337} & \rankSecond{182.433} & \rankFirst{0.112} & \rankFirst{92.543} \\
\bottomrule
\end{tabular}}
\caption{Downstream generation on UCF-101 and SkyTimelapse. Baseline results follow their respective papers and native generator settings; MotionStrata uses the LatteMotion-style conditional generator described in Appendix~S5.}
\label{tab:downstream}
\end{table}

\begin{table}[t]
\centering
\small
\setlength{\tabcolsep}{1.8pt}
\resizebox{\columnwidth}{!}{%
\begin{tabular}{lcccc}
\toprule
Method & Input & Enc.$\downarrow$ & Dec.$\downarrow$ & Peak mem.$\downarrow$ \\
\midrule
WF-VAE & $17{\times}256^2$ & \rankSecond{$18.97{\pm}0.06$} & \rankSecond{$57.77{\pm}0.78$} & \rankSecond{2.797} \\
Reducio ($f_t{=}2$) & $16{\times}256^2$ & $70.26{\pm}0.05$ & $175.65{\pm}0.36$ & \rankThird{6.150} \\
Reducio ($f_t{=}4$) & $16{\times}256^2$ & $69.41{\pm}0.10$ & \rankThird{$174.66{\pm}0.27$} & 6.167 \\
VidTwin & $16{\times}224^2$ & \rankFirst{$15.33{\pm}0.44$} & \rankFirst{$12.28{\pm}0.11$} & \rankFirst{1.363} \\
MotionStrata (5 NFE) & $16{\times}256^2$ & \rankThird{$47.83{\pm}0.21$} & $296.82{\pm}0.29$ & 6.952 \\
MotionStrata (20 NFE) & $16{\times}256^2$ & \rankThird{$47.83{\pm}0.21$} & $965.08{\pm}2.18$ & 6.952 \\
\midrule
FlowMo-Lo (image) & $16{\times}256^2$ frames & $275.04{\pm}2.43$ & $26089.46{\pm}8.17$ & 10.083 \\
\bottomrule
\end{tabular}}
\caption{Measured inference efficiency on one NVIDIA L20X in FP32. Video checkpoints use batch 1 and their supported native shapes; WF-VAE requires $4k{+}1$ frames and VidTwin uses $224^2$. FlowMo-Lo instead processes 16 independent images as one batch. Enc./Dec. are milliseconds (mean$\pm$std over 30 runs after 5 warm-ups).}
\label{tab:efficiency_protocol}
\end{table}

\begin{table}[t]
\centering
\small
\setlength{\tabcolsep}{3.4pt}
\begin{tabular}{rcccc}
\toprule
NFE & Dec. ms$\downarrow$ & $\Delta$PSNR$\uparrow$ & LPIPS rel.$\downarrow$ & rFVD rel.$\downarrow$ \\
\midrule
1 & \rankFirst{$117.94{\pm}0.26$} & \rankSecond{$+0.852$} & 1.451 & 4.077 \\
2 & \rankSecond{$162.69{\pm}0.24$} & \rankFirst{$+0.886$} & 1.266 & 2.815 \\
5 & \rankThird{$296.82{\pm}0.29$} & \rankThird{$+0.534$} & \rankThird{1.069} & \rankThird{1.453} \\
10 & $519.71{\pm}0.40$ & $+0.232$ & \rankSecond{1.011} & \rankSecond{1.083} \\
20 & $965.08{\pm}2.18$ & $0.000$ & \rankFirst{1.000} & \rankFirst{1.000} \\
\bottomrule
\end{tabular}
\caption{MotionStrata quality--latency trade-off on a fixed 2,048-clip WebVid manifest. $\Delta$PSNR is measured relative to 20 NFE; LPIPS and rFVD are divided by their 20-NFE values. Latency follows the L20X protocol in Supp. Sec.~6.}
\label{tab:nfe_curve}
\end{table}

\noindent\textbf{Decoding efficiency.}
Table~\ref{tab:efficiency_protocol} measures latency and memory on one L20X; FlowMo-Lo~\cite{sargent2025flow} is a separate 16-image reference rather than a video model. Table~\ref{tab:nfe_curve} sweeps 1--20 NFE: more steps improve LPIPS and rFVD, while 5 NFE reduces decoder latency from 965.08 to 296.82 ms. Appendix~S6 gives the full protocol.

\noindent\textbf{Seed stability.}
The main metrics use one fixed draw per clip. Separately, we hold the encoded conditions fixed and vary only the flow initialization over five seeds. The resulting variances are $3.69{\times}10^{-6}$ for LPIPS, $2.61{\times}10^{-5}$ for CLIP-to-GT similarity, and $7.13{\times}10^{-3}$ for flow difference, quantifying decoder sensitivity to initialization under fixed encoded conditions (Appendix~S1).

\section{Limitations}
\label{sec:limitations}

MotionStrata is trained on short clips; longer and native-resolution inference uses anchor refresh and tiling. Generation is evaluated only for feasibility, with quality and decoding speed left for future work.

\section{Conclusion \& limitation}
MotionStrata organizes a compact motion code as temporally compressed Global latents and frame-aligned Detailed latents. Matched controls show that frequency-guided routing, asymmetric temporal support, and staged training improve reconstruction under a fixed motion budget. Additional stress tests and downstream experiments characterize its operating range beyond the primary in-domain comparison. These results establish motion organization as an important design axis for compact video autoencoding.

\bibliography{main}

\clearpage
\appendix
\setcounter{section}{0}
\renewcommand{\thesection}{S\arabic{section}}
\section*{Supplementary Material}
\noindent\textbf{Supplementary overview.}
\begin{center}
\small
\setlength{\tabcolsep}{4pt}
\begin{tabular}{@{}cl@{}}
\toprule
Section & Contents \\
\midrule
\ref{supp:sec:evaluation} & Datasets, manifests, metrics, and stress-test protocols \\
\ref{supp:sec:implementation} & Reproduction details, matched controls, and accounting \\
\ref{supp:sec:diagnostics} & Temporal-spectrum and stream-removal diagnostics \\
\ref{supp:sec:qualitative} & Additional reconstruction results \\
\ref{supp:sec:generation} & Downstream generation protocol and cost \\
\ref{supp:sec:efficiency} & Timing protocol and FlowMo scope \\
\bottomrule
\end{tabular}
\end{center}

\section{Evaluation Protocol}
\label{supp:sec:evaluation}

Table~\ref{tab:evaluation_protocol} summarizes the role of each dataset. Reconstruction evaluations use fixed manifests and identical preprocessing across methods within each controlled comparison. Contextual downstream baselines retain their native training and evaluation protocols.

\begin{center}
\centering
\small
\setlength{\tabcolsep}{3.0pt}
\resizebox{\columnwidth}{!}{%
\begin{tabular}{lrrrll}
\toprule
Dataset & Clips & Frames & FPS & Resolution & Use \\
\midrule
WebVid-10M & 2,048 & 16 & 8 & $256^2$ & in-domain reconstruction \\
WebVid-10M (long) & 500 & 16/32/64 & 8 & $256^2$ & length stress test \\
UCF-101 & 2,000 & 16 & 8 & $256^2$ & zero-shot reconstruction / downstream generation \\
RealEstate10K & 2,000 & 16 & 8 & $256^2$ & zero-shot reconstruction \\
RealEstate10K (native) & 500 & 16 & 8 & $832\!\times\!480$ & resolution stress test \\
SkyTimelapse & -- & 16 & 8 & $256^2$ & downstream generation \\
\bottomrule
\end{tabular}}
\captionof{table}{Dataset roles and evaluation manifests. Counts refer to reconstruction or stress-test manifests; downstream generation follows its dataset-specific protocol.}
\label{tab:evaluation_protocol}
\end{center}

\noindent\textbf{Long-video and native-resolution tests.}
For the length stress test, we select WebVid clips that support 65 sampled frames at 8 fps. Each long clip is divided into consecutive 16-frame segments, and the first frame of every segment is encoded as its content anchor. The 16-, 32-, and 64-frame conditions therefore use one, two, and four anchors, respectively, while sharing the same sampled trajectory. For the resolution stress test, we retain native $832\!\times\!480$ RealEstate10K clips and blend overlapping $256^2$ tiles with a 128-pixel overlap. Both tests use frame-wise VAE reconstructions as reference, isolating degradation from the compact motion interface rather than the image-VAE reconstruction ceiling.

\noindent\textbf{Metric implementation.}
Pixel metrics are averaged over all reconstructed frames. rFID uses 2,048-dimensional Inception-V3 frame features, and rFVD uses 400-dimensional I3D video features. The temporal-spectrum and stream-removal diagnostics below use fixed seed-42.

\noindent\textbf{Randomness and run counts.}
We use a fixed random seed of 42 for reconstruction evaluation and generate one reconstruction per clip. Metrics are aggregated over the fixed manifest without best-of-$N$ selection or averaging multiple decoding runs. Each model or ablation row reports one training run and one selected checkpoint. The NFE latency study uses 30 timed repetitions after five warm-ups. For the separate stability diagnostic, we hold the encoded conditions fixed and vary only the flow initialization over seeds 0--4, obtaining variances of $3.69{\times}10^{-6}$, $2.61{\times}10^{-5}$, and $7.13{\times}10^{-3}$ for LPIPS, CLIP-to-GT similarity, and flow difference.

\section{Implementation and Accounting}
\label{supp:sec:implementation}

\noindent\textbf{Representation contract.}
The frozen base image VAE is the Diffusers \texttt{stabilityai/sd-vae-ft-mse} AutoencoderKL checkpoint. Inputs are normalized to $[-1,1]$; one posterior sample per frame is multiplied by 0.18215 before entering MotionStrata, and reconstructed latents are divided by 0.18215 before VAE decoding. It maps each $256\!\times\!256$ frame independently to a $32\!\times\!32\!\times\!4$ latent. The content anchor is the sampled first-frame latent and therefore has no temporal receptive field. In temporal--query--channel order, MotionStrata uses Global and Detailed tensors of size $8\!\times\!8\!\times\!8$ and $16\!\times\!8\!\times\!32$, respectively. The Detailed level preserves latent-frame alignment, while the Global level uses a shorter temporal query grid.

\noindent\textbf{FFT implementation.}
We apply \texttt{fftn} independently to each latent channel over the temporal, height, and width axes, followed by \texttt{fftshift} on the same three axes. Frequencies are indexed relative to the centered zero-frequency bin and normalized by the Nyquist frequency of each axis. The hard cuboid $P_k$ retains entries whose normalized absolute index is at most $k$ on every routed axis; its exact complement $1-P_k$ defines the residual branch. We preserve the complex coefficients and phase, apply \texttt{ifftshift} and \texttt{ifftn}, and retain the real part of the inverse transform. Temporal-only and spatial-only controls apply the same construction to the named axes and pass all frequencies on the remaining axes.

\noindent\textbf{Architecture and conditioning.}
Both motion encoders contain eight Transformer blocks with eight heads and width 512; each encoder has approximately 25M parameters. Base-latent features are embedded with $2\times2$ patches; fixed 2D sine--cosine spatial embeddings are combined with fixed 1D temporal/query embeddings. The conditional flow decoder contains 16 Transformer blocks and approximately 640M parameters, using the same $2\times2$ latent patch size. Global and Detailed tokens are linearly projected to the decoder width and injected through attention in every decoder block before the spatial--temporal state update. At inference, a uniform-grid explicit Euler solver integrates the rectified-flow velocity from noise to data; the default is 20 function evaluations.

\noindent\textbf{Optimization and software.}
Stage 1 and Stage 2 use 20k and 80k updates, requiring approximately 200 and 750 GPU hours, respectively. We use Adam with $(\beta_1,\beta_2)=(0.9,0.99)$, learning rate $10^{-4}$, weight decay $10^{-4}$, 5k warm-up steps followed by cosine annealing, a global batch size of 8, and $\lambda_{\mathrm{KL}}=0.001$. Training uses FP32 on eight 80GB NVIDIA A800 GPUs; evaluation was conducted on an NVIDIA RTX 4090. The flow-matching scheduler represents training time with 1,000 discrete levels, and inference uses 20 Euler steps by default. Classifier-free guidance drops conditioning for 10\% of training samples and uses guidance scale 5 at inference. The recorded reproduction environment uses Python 3.10, PyTorch 2.2.0 with CUDA 12.1, and Diffusers 0.30.0.

\noindent\textbf{Matched-control definitions.}
\emph{Flat} uses one homogeneous motion stream with the same total scalar dimensionality. \emph{Random split} uses MotionStrata's two branch shapes but assigns input latent tokens to the branches with a fixed random partition. \emph{Learned split} replaces that partition with a learned two-way router. \emph{Full-band} gives both asymmetric branches the complete base latent without spectral routing. \emph{Symmetric grids} retain 3D routing but give both branches the frame-aligned query grid. \emph{Temporal FFT} routes by temporal frequency while passing all spatial frequencies, and \emph{Spatial FFT} routes by spatial frequency while passing all temporal frequencies. Unless explicitly varied, these controls share the content anchor, total motion dimensionality, encoder depth, flow decoder, sampler, and 100k-update budget.

\begin{center}
\centering
\small
\setlength{\tabcolsep}{4.0pt}
\begin{tabular}{lccc}
\toprule
Tensor & Shape & Scalars & Temporal support \\
\midrule
Base latent $z$ & $16\times32\times32\times4$ & 65,536 & 16 frames \\
Content $c_0$ & $1\times32\times32\times4$ & 4,096 & first frame \\
Global $u_g$ & $8\times8\times8$ & 512 & 8 positions \\
Detailed $u_d$ & $16\times8\times32$ & 4,096 & 16 positions \\
\bottomrule
\end{tabular}
\captionof{table}{Tensor contract for the default 16-frame model. Motion shapes follow temporal--query--channel order.}
\label{tab:supp_tensor_contract}
\end{center}

\begin{center}
\centering
\small
\setlength{\tabcolsep}{4pt}
\begin{tabular}{lrrrr}
\toprule
Variant & $|u_g|$ & $|u_d|$ & $r_{\mathrm{motion}}$ & $r_{\mathrm{total}}$ \\
\midrule
MotionStrata (S) & 256 & 2,048 & \rankFirst{0.073\%} & \rankFirst{0.203\%} \\
MotionStrata & 512 & 4,096 & \rankSecond{0.146\%} & \rankSecond{0.277\%} \\
MotionStrata (L) & 512 & 8,192 & \rankThird{0.277\%} & \rankThird{0.407\%} \\
\bottomrule
\end{tabular}
\captionof{table}{Motion-only and total-conditioning accounting for a 16-frame $256\!\times\!256$ RGB clip. The first-frame content anchor contributes 4,096 scalars.}
\label{tab:supp_accounting}
\end{center}

Here $3FHW=3{,}145{,}728$. The motion ratio counts only $u_g$ and $u_d$, whereas the total ratio also counts the first-frame content anchor. Both are scalar-dimensionality ratios rather than entropy-coded bitrates. For external systems, the main reconstruction table reports total conditioning only when every reconstruction input can be identified from the released interface.

\begin{figure}[t]
  \centering
  \includegraphics[width=0.94\columnwidth]{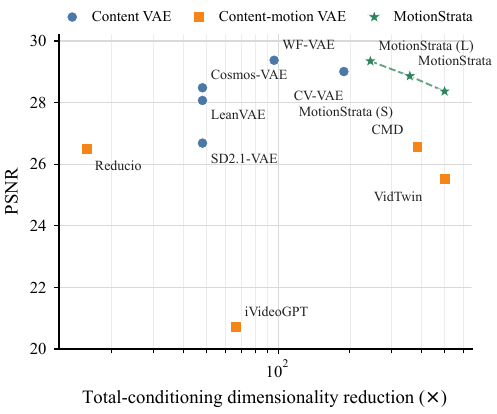}
  \caption{Reconstruction quality versus total-conditioning dimensionality on 16-frame WebVid clips. Architectural differences remain, so this figure provides cross-system context rather than a controlled causal comparison.}
  \label{fig:frontier}
\end{figure}

\section{Supporting Diagnostics}
\label{supp:sec:diagnostics}

\noindent\textbf{Temporal spectral concentration.}
We analyze 2,048 WebVid validation clips using the same 16-frame, 8-fps sampling protocol as reconstruction. Frames are encoded with the frozen base VAE posterior mean. We subtract the temporal mean independently at every latent channel and spatial position, then apply a one-sided temporal FFT. Bins 1--4 and 5--8 cover the lower and upper halves of the non-DC spectrum, respectively.

\begin{center}
\centering
\small
\setlength{\tabcolsep}{4.0pt}
\resizebox{\columnwidth}{!}{%
\begin{tabular}{lcc}
\toprule
Temporal band & One-sided bins & Dynamic-energy share (95\% CI) \\
\midrule
Low & 1--4 & 79.99\% [79.58, 80.40] \\
High & 5--8 & 20.01\% [19.60, 20.42] \\
\bottomrule
\end{tabular}}
\captionof{table}{Temporal spectral concentration of frozen base latents on 2,048 WebVid validation clips. For 16-frame latents, bins 1--4 and 5--8 cover the lower and upper halves of the non-DC one-sided spectrum. Confidence intervals are obtained by clip-level bootstrap.}
\label{tab:supp_temporal_spectrum}
\end{center}

The lower temporal band contains approximately four times the dynamic energy of the upper band. This model-independent statistic motivates using different temporal support for broad evolution and weaker, faster-changing evidence; it does not by itself establish a semantic separation between the two learned motion groups.

\noindent\textbf{Stream-removal protocol.}
For the main-paper stream-removal analysis, we use a fixed 500-clip WebVid manifest. Each clip is encoded once, and the content anchor, ODE schedule, and clip-wise initial decoder noise are held fixed. We then zero $u_g$ or $u_d$ after encoding without retraining. For a diagnostic band $b$, normalized spectral error is
\begin{equation}
 E_b=\left(
 \frac{\lVert b\odot\mathcal{F}(\hat z-z)\rVert_2^2}
 {\lVert b\odot\mathcal{F}(z)\rVert_2^2}
 \right)^{1/2}.
 \label{eq:supp_band_error}
\end{equation}
Dense adjacent-frame flow is estimated with RAFT. We fit a homography independently to reconstructed and target flow, report EPE between the fitted dense flows, and compute residual-flow EPE after subtracting each fitted component. The experiment measures asymmetric sensitivity after removing one level; it does not assume that the two levels are independent or mutually exclusive.

\section{Additional Qualitative Results}
\label{supp:sec:qualitative}

\begin{figure}[t]
  \centering
  \includegraphics[width=\columnwidth]{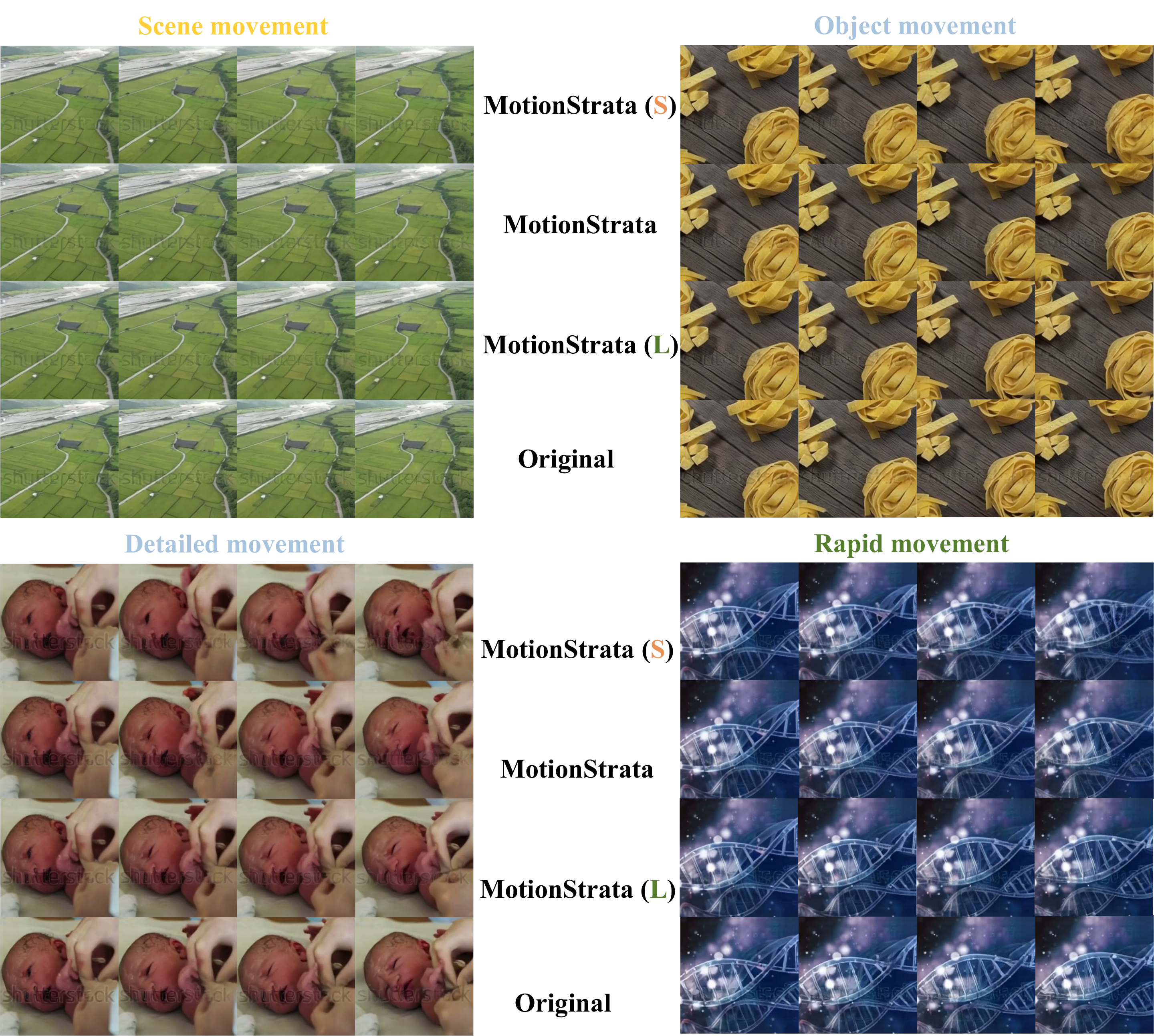}
  \caption{Qualitative reconstruction across MotionStrata motion budgets. Additional capacity primarily improves fast local changes and fine structures.}
  \label{fig:supp_motion_capacity}
\end{figure}

\begin{figure*}[t]
  \centering
  \includegraphics[width=0.92\textwidth]{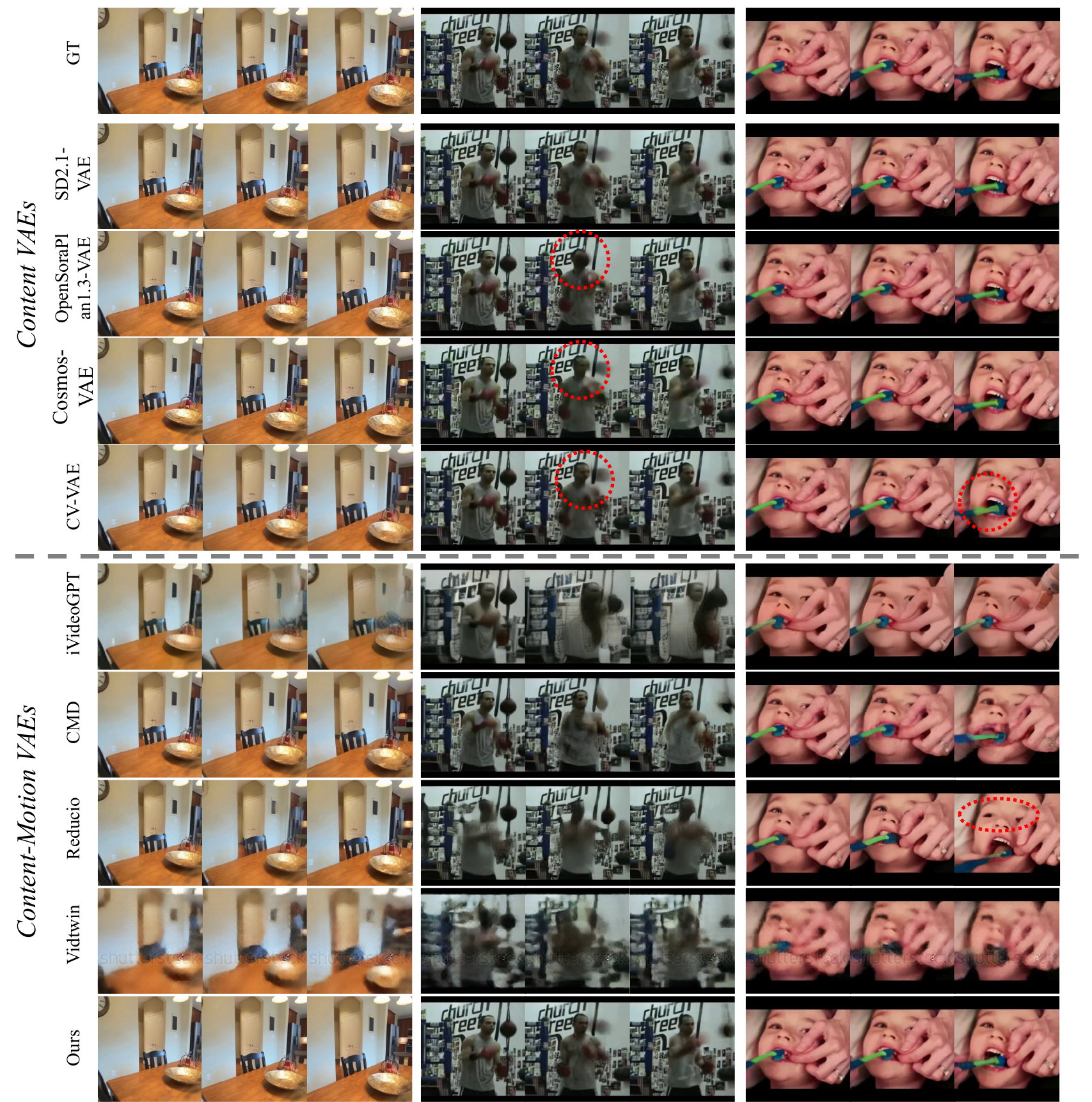}
  \caption{Additional reconstruction examples under the default 16-frame WebVid protocol.}
  \label{fig:supp_reconstruction}
\end{figure*}

Figure~\ref{fig:supp_reconstruction} extends the main qualitative comparison with additional subjects and motion patterns. Figure~\ref{fig:supp_motion_capacity} shows how the small, default, and large MotionStrata variants change reconstruction as the motion budget grows.

\section{Downstream Generation Protocol}
\label{supp:sec:generation}

\noindent\textbf{Dataset construction.}
For WebVid, we use the provided text annotations. For UCF-101 and SkyTimelapse clips with weak or missing text, we generate concise clip-level captions describing the visible subject, action, and scene. SkyTimelapse retains one segment per source video to reduce near duplicates. We consider category-driven generation and text-plus-first-frame conditioned generation.

\noindent\textbf{Motion representation.}
The frozen MotionStrata encoder extracts $u_g\in\mathbb{R}^{f_g\times n_g\times c_g}$ and $u_d\in\mathbb{R}^{f_d\times n_d\times c_d}$. The two groups are projected to a shared channel width $c_m$, normalized, temporally aligned, and concatenated along the token axis to form the diffusion target
\begin{equation}
 m\in\mathbb{R}^{f\times(n_g+n_d)\times c_m}.
 \label{eq:supp_generation_target}
\end{equation}

\noindent\textbf{Conditional generator.}
For text-plus-first-frame generation, we use a DiT-based conditional motion generator following the LatteMotion design. Text tokens, content tokens obtained by patchifying the first-frame latent, and noised motion tokens form one joint Transformer sequence. Text provides global semantic guidance, while the first frame is supplied through global conditioning and content-prefix tokens. Starting from Gaussian motion tokens, the model predicts a hierarchy consistent with both conditions; the frozen MotionStrata decoder then reconstructs the video.

\begin{figure}[t]
  \centering
  \includegraphics[width=0.9\columnwidth]{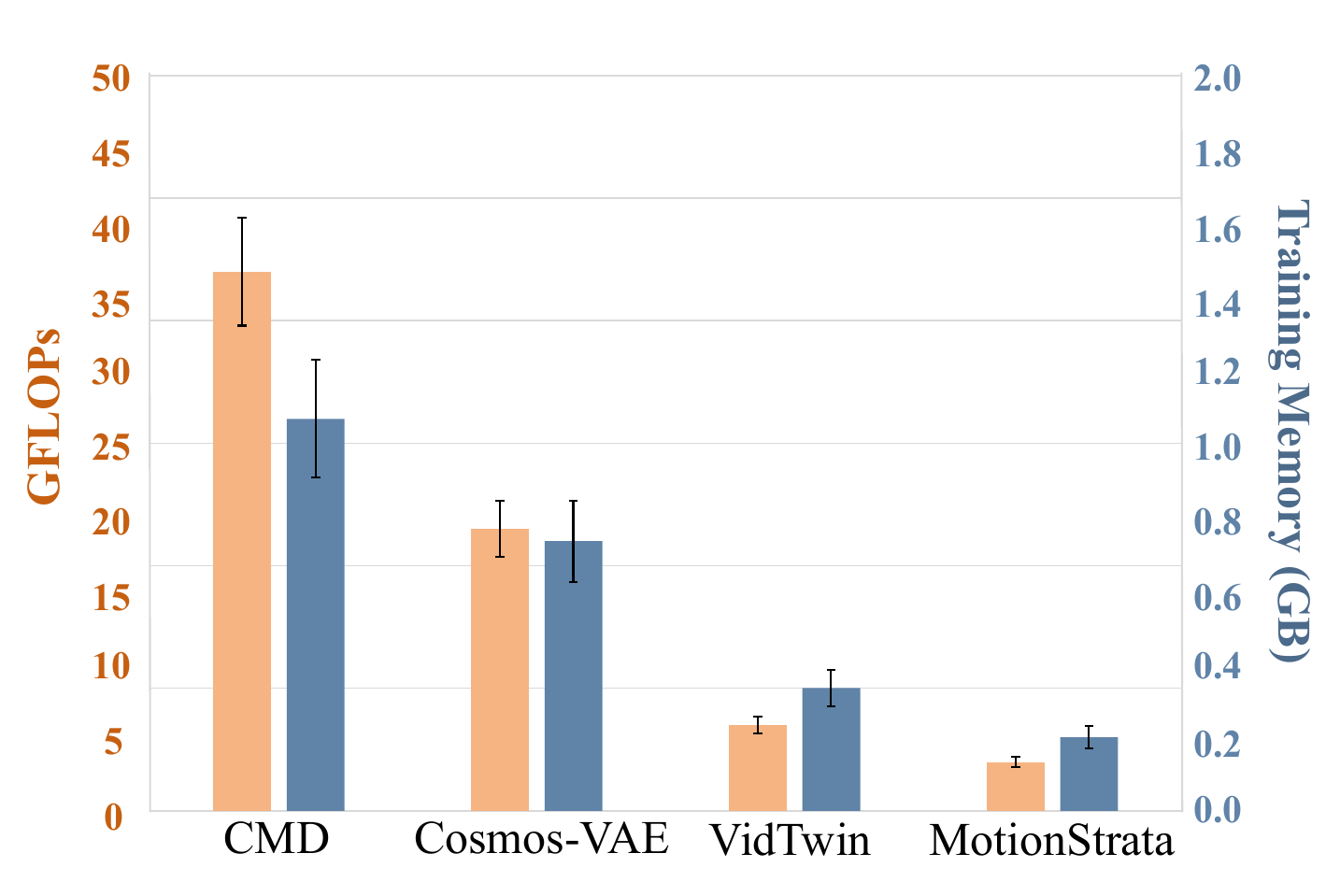}
  \caption{Generator GFLOPs and training memory under a standardized DiT with eight layers, eight heads, hidden width 512, and FFN width 2048. Values are averaged over ten dataset splits; error bars denote standard deviation.}
  \label{fig:supp_generator_efficiency}
\end{figure}

\noindent\textbf{Training objective.}
Let $m$ be the clean motion representation and $\epsilon\sim\mathcal{N}(0,I)$. The generator predicts the velocity $v$ of the forward diffusion process using
\begin{equation}
 \mathcal{L}_{\mathrm{diff}}=\lVert\hat v-v\rVert_2^2.
 \label{eq:supp_generation_diff}
\end{equation}
We additionally supervise the clean prediction and its temporal differences for both motion groups. With $s\in\{g,d\}$ and Smooth-L1 distance $\ell$,
\begin{align}
 \mathcal{L}_{x_0}
 &=\sum_s\lambda_s\ell(\hat m_0^{(s)},m^{(s)}),\\
 \mathcal{L}_{\Delta}
 &=\sum_s\lambda_s\ell(\Delta\hat m_0^{(s)},\Delta m^{(s)}),\\
 \mathcal{L}_{\mathrm{acc}}
 &=\sum_s\lambda_s\ell(\Delta^2\hat m_0^{(s)},\Delta^2m^{(s)}),
 \label{eq:supp_generation_temporal}
\end{align}
where $\Delta m_t=m_t-m_{t-1}$ and $\Delta^2m_t=m_t-2m_{t-1}+m_{t-2}$. Some settings also apply the first- and second-order terms after decoding the predicted motion. The complete objective is
\begin{equation}
\begin{aligned}
 \mathcal{L}_{\mathrm{gen}}={}&\mathcal{L}_{\mathrm{diff}}
 +\lambda_{x_0}\mathcal{L}_{x_0}
 +\lambda_{\Delta}\mathcal{L}_{\Delta}
 +\lambda_{\mathrm{acc}}\mathcal{L}_{\mathrm{acc}}\\
 &+\lambda_{\Delta}^{\mathrm{dec}}\mathcal{L}_{\Delta}^{\mathrm{dec}}
 +\lambda_{\mathrm{acc}}^{\mathrm{dec}}\mathcal{L}_{\mathrm{acc}}^{\mathrm{dec}}.
\end{aligned}
\label{eq:supp_generation_total}
\end{equation}

\noindent\textbf{Optimization.}
We use Adam with $(\beta_1,\beta_2)=(0.9,0.999)$ and a warm-up learning-rate schedule. Training uses 16-frame, $256^2$ clips on eight A800 GPUs. At inference, the predicted motion tokens are passed to the frozen MotionStrata decoder. All baseline numbers in the main-paper downstream table are quoted from the corresponding papers under their native generator settings.

\section{Efficiency Protocol}
\label{supp:sec:efficiency}

The main-paper efficiency table reports newly measured wall-clock latency on one NVIDIA L20X with FP32 inference and TF32 enabled. Each result uses five warm-up iterations followed by 30 synchronized CUDA-event measurements; data loading and host-to-device transfer are excluded. Peak memory is the maximum allocated GPU memory over the complete encode--decode path. Public video checkpoints use batch size 1 and their supported native shapes, which are stated explicitly in the table.

FlowMo-Lo is measured with its official checkpoint and eager inference using 25 NFE, classifier-free guidance 1.5, and the \texttt{pow\_0.25} schedule. To match the number of frames processed by the video benchmark, its formal row batches 16 independent $256^2$ images, yielding $275.04{\pm}2.43$ ms encoding, $26089.46{\pm}8.17$ ms decoding, and 10.083 GiB peak allocated memory. Because FlowMo has no temporal encoder or video interface, this row is reported separately and is not ranked against the video autoencoders.

\clearpage

\end{document}